\documentclass[runningheads]{llncs}
\usepackage{graphicx}
\usepackage{amsmath,amssymb} 
\usepackage{color}
\usepackage[width=122mm,left=12mm,paperwidth=146mm,height=193mm,top=12mm,paperheight=217mm]{geometry}
\usepackage[utf8]{inputenc}
\usepackage{pgf,tikz}
\usepackage{ifthen}
\usepackage{mathrsfs, stmaryrd}
\usepackage{graphbox}
\usepackage{graphics}
\usetikzlibrary{arrows.meta,calc,decorations.markings,math,arrows.meta}
\usepackage[table]{colortbl}
\usepackage{ctable}
\usepackage{cite}

\mainmatter
\def\ECCV18SubNumber{18}  

\title{Learning 
structure-from-motion
from motion}
\author{Clément Pinard\inst{1,2} \and Laure Chevalley\inst{2} \and Antoine Manzanera\inst{1} \and David Filliat\inst{1}
}

\institute{
ENSTA ParisTech \\
Computer Science and System Engineering Department\\
Palaiseau, France\\
\email{\{clement.pinard, antoine.manzanera, david.filliat\}@ensta-parsitech.fr}
\and
Parrot,
Paris, France\\
\email{laure.chevalley@parrot.com}
}

\def\eg{\emph{e.g.}\:}

\def\etal{\emph{et al}\:}
\def\ie{\emph{i.e.}\:}

\authorrunning{Clément Pinard \etal}

\begin{document}

\maketitle

\begin{abstract}
This work is based on a questioning of the quality metrics used by deep neural networks performing depth prediction from a single image, and then of the usability
of recently published works on unsupervised learning of depth from videos. These works are all predicting depth from a single image, thus it is only known up to an undetermined scale factor,
which is not sufficient for practical 
use cases that need an absolute depth map, i.e. the determination of the scaling factor. 
To overcome these limitations, we propose to learn in the same unsupervised manner a depth map inference system from monocular videos that takes a pair of images as input. This algorithm
actually
learns 
structure-from-motion {\em from motion}, and not only structure from context appearance.
The scale factor issue is explicitly treated, and the absolute 
depth map
can be estimated from camera displacement magnitude, which can be easily measured from cheap external sensors. Our solution is also 
much more robust with respect to
domain variation and adaptation via 
fine tuning,
because it does not rely entirely on depth from context. Two 
use cases
are considered, 
unstabilized moving camera videos,
and stabilized ones. This choice is motivated by the UAV (for Unmanned Aerial Vehicle)
use case
that generally provides reliable orientation measurement. We provide a set of experiments 
showing that, used
in real conditions where only speed can be known, our network outperforms competitors for most depth quality measures. Results are given on the well known KITTI dataset \cite{geiger2013vision}, which provides robust stabilization for our second 
use case,
but 
also
contains moving scenes which are 
very typical of the in-car road context. We then present results on a synthetic dataset that we believe to be more representative of typical UAV scenes. Lastly, we present two domain adaptation use cases showing superior robustness of our method compared to single view depth algorithms, 
which indicates that
it is better suited for 
highly variable visual contexts.
\end{abstract}

\section{Introduction}
\label{sec:intro}
Scene understanding from vision, in particular depth estimation, is a core problem for autonomous vehicles.

One could train a system for depth from vision with supervised learning on an offline dataset which features explicit depth measurement, such as KITTI \cite{geiger2013vision}, but even setting up such recording devices can be costly and time demanding, which can limit the amount of data the system can be trained on.

As a consequence, in this paper we are specifically interested in \textbf{unsupervised} learning of depth from images using machine learning optimization techniques.

Training to infer the depth of a scene and one's ego-motion is a problem for which recent work has been successfully done with no supervision, leveraging uncalibrated data solely from RGB cameras, but to our knowledge, all of them infer depth from a single image
\cite{zhou2017unsupervised,yin2018geonet,monodepth17,Vijayanarasimhan17,Mahjourian}. On the contrary, our methods tries to deduce it from multiple frames, using motion instead of context.

UAV navigation, which is one of our favorite use cases, is very specific compared to other ego motion videos. Its two main characteristics are the availability of orientation and the high variability of the visual context:
 
\begin{itemize}
\item An UAV relies on inertial data to maintain its position and the current market for UAVs allows to get high quality video stabilization even for consumer products. As such, orientation of any frame can be assumed to be well estimated.
\item Compared to videos acquired from any other vehicle,
UAV scenes are very heterogeneous. Unlike a camera fixed to a car, altitude can vary widely and quickly, along with velocity, orientation, and scene 
visual layout,
which context can be hard to figure out with only one frame.
\end{itemize}

Hence, we propose an unsupervised scene geometry learning algorithm that aims at inferring a depth map from a sequence of images.
Our algorithm works with stabilized {\em and} unstabilized videos, the latter
requiring in addition to the depth estimator network an orientation estimator that can bring back digital stabilization.

Our algorithm outputs a depth map assuming a constant displacement, this solves the scale factor incertitude simply by knowing ego-motion speed.
This allows a straightforward real conditions depth inference process for any camera with a known speed such as the one of the supporting cars or UAVs.

\section{Related Work}

First works trying to compute a depth map from images using machine learning can be found as early as 2009 \cite{4531745}. Whether from multiple frames or a single frame, these techniques have shown great generalization capabilities, especially using end-to-end learning methods such as convolutional neural networks.

\subsection{Supervised Depth Networks}
Most studied problems for supervised depth learning use a stereo rig along with corresponding disparity \cite{zbontar2016stereo, Kendall2017EndtoEndLO}, thanks to dedicated datasets \cite{geiger2013vision, MIFDB16}. For unconstrained monocular sequences, DepthNet and DeMoN \cite{isprs-annals-IV-2-W3-67-2017,UZUMIDB17} are probably the works closest to ours.
Using depth supervision, these networks aim to compute a depth map with a pair of images from a monocular video.

The first network explicitly assumes a fixed displacement magnitude and a stabilized video, and only outputs depth, while the second one also outputs a pose, from which translational component is trained to be of constant magnitude.
Both methods easily solve the scale factor problem when the camera speed is known. Our main goal here is to achieve the same operation, but with unsupervised training.

\subsection{Unsupervised Depth Learning}
Most recent works on unsupervised training networks for computing depth maps use differentiable bilinear warping techniques, first introduced in \cite{jaderberg2015spatial}. The main idea is trying to match two frames using a depth map and a displacement. The new loss function to be minimized is the photometric error between the reference frame and the projected one. Depth is then indirectly optimized. Although sensitive to errors coming from occlusions, non Lambertian surfaces and moving objects, this optimization shows great potential, especially when considering how little calibrated data is needed.

For instance \cite{monodepth17, garg2016unsupervised, xie2016deep3d} use stereo views and try to reconstruct one frame from the other. This particular use case for depth training allows to always consider the same displacement and rigid scenes since both images are captured at the same time and their relative poses are always the same. However, it constraints the training set to stereo rigs, which are not as easy to set up as a monocular camera.

When trying to estimate both depth and movement, \cite{zhou2017unsupervised, yin2018geonet, Mahjourian, DBLP:journals/corr/abs-1805-09806} also achieved decent results on completely unconstrained ego-motion video. One can note that some methods \cite{zhou2017unsupervised} are assuming rigid scenes although the training set does not always conform to this assumption. The other ones try to 
do without this assumption by computing
a residual optical flow 
to resolve
the uncertainty from moving objects.

\cite{Vijayanarasimhan17} explicitly considered non rigid scenes by trying to estimate multiple objects movements in the scene, to begin with the motion of the camera itself, which allowed them to deduce a flow map, along with the depth map. The reader is 
referred to
the work of Zhou \etal \cite{zhou2017unsupervised} for a more complete vision of the field, as all other works 
are actually built
on this fundamental basis.

\section{Single Frame Prediction 
{\em vs}
Reality}

\label{section:critic}
As already mentioned, in the current state of the art 
of
learning from monocular footage, depth is always inferred by the network using a single image. Indeed, it is mentioned in \cite{zhou2017unsupervised} that feeding multiple frames to a network did not yield better results. This may be due to the fact that for particular scenarii with very 
typical geometric and photometric contexts
such as in-car view of the road, depth seems easier to get from the 
visual layout
than from the motion.

Because of the single frame context, current depth quality measurements, originally introduced by Eigen \etal \cite{eigen2014depth}, expects a relative depth map up to a scale factor. This is problematic, as they completely ignore the scale factor uncertainty, and thus rely on 
estimating the scale factor as
the ratio of the medians of network's output and groundtruth. This is then representative to an ideal use case where the median of an unknown depth map has to be available, which is clearly unrealistic.

One can try to overcome these limitations by figuring out the scale factor with several solutions:
\begin{itemize}
\item Measuring depth, at least in one point, with additional sensor such as LiDAR, 
Time-of-Flight
or stereo cameras. This is not a trivial solution and it needs integration, as 
well
as precise calibration.

\item Assuming depth consistency across training and testing dataset. This can be particularly useful in datasets like KITTI \cite{geiger2013vision}, where the camera is always at the same height and looking at the floor 
from
the same angle, but it 
is irrelevant on a dataset with high pose variability, 
\eg UAV videos, and such assumptions will fail.
\end{itemize}

Thankfully, those techniques 
do
not only predict depth, but also ego-motion. An other network tries to compute poses of frames in a sequence. Considering that the 
uncertainties
about depth in one hand and pose in the other hand are consistent, the scale factors for depth and pose are theoretically the same. 
As a consequence, depth scale factor can be determined from the ratio between translation estimation, and actual translation measurement, which is already available on cars and UAVs.

A new quality measurement can then be designed, by slightly modifying the already prevalent ones. This new measurement is not relative anymore, it computes actual depth errors, and is more representative of a real application case. This will be denoted in tables \ref{table:KITTI}, \ref{table:StillBox} and \ref{table:KITTI_reversed} by the scale factor column. When using standard relative depth measurement, the indication $GT$ (for \emph{Ground Truth}) is used, when using translation magnitude, the letter $P$ (for \emph{Pose}) is used.

\section{Approach}

\begin{figure}\centering
\begin{tikzpicture}
  \node[yslant=0.2, scale=0.2, opacity=0.5]  at (0.2,0) {\includegraphics{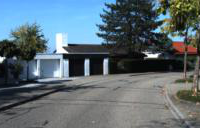}};
  \node[yslant=0.2, scale=0.2]  at (1.1,0) {\includegraphics{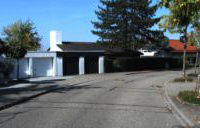}};
  \node at (1.3,1) {$I_r$};
  \node at (3,1) {$I_t$};
  \node[yslant=0.2, scale=0.2, opacity=0.5]  at (2,0) {\includegraphics{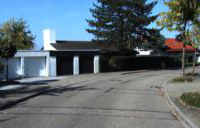}};
  \node[yslant=0.2, scale=0.2]  at (2.9,0) {\includegraphics{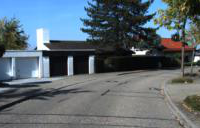}};
  \node[yslant=0.2, scale=0.2, opacity=0.5]  at (3.8,0) {\includegraphics{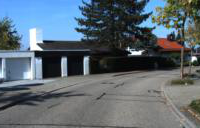}};
  \node (depth) at (7,-2) {\includegraphics[scale=0.15]{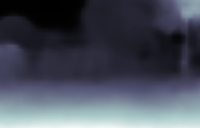}};
  
  \node (stab) [scale=0.15]  at (3.2,-2) {\includegraphics{figure/KITTI/unstab3}};
  \node at (3.2,-2.6) {$I^\textsc{stab}_r$};
  \draw (-1.5,-2.5) node[below] {$PoseNet$};
  \draw (5,-2.5) node[below] {$DepthNet$};
  \node (inverse_rot) [rectangle,draw=black,align=center,fill=blue!30,text width=30,scale=0.7] at (1,-2){Inverse rot};
  \node (inverse warp) [rectangle,draw=black,align=center,fill=blue!30,text width=30,scale=0.7] at (8.5,-2){Inverse Warp};
  \node (loss) [rectangle,draw=black,align=center,fill=green!30,text width=25,scale=0.7] at (9.8,-2){Photo loss};
  
  \draw [>=stealth,->] (-1,-0) -| ++(-1.5,-2) -- ++(0.5,0);
  \draw [>=stealth,->] (4.7,0) -| (inverse warp);
  \draw [>=stealth,->] (3,0.5) -- ++(0,0.15) -| (loss);
  \draw [>=stealth,->] (0,-2) -- ++(0,-1.2) -| (inverse warp) node [near start,below]{$\forall i, t_i^{\textsc{norm}} = t_i\frac{D_0}{\epsilon + \Vert t_r\Vert}$};
  \draw [>=stealth,->] (1,-0.5) -- (inverse_rot);
  \draw [>=stealth,->] (stab) -- ++(1.4,0);
  \draw [>=stealth,->] (inverse_rot) -- (stab);
  \draw [>=stealth,->] (-1.,-2) -- (inverse_rot) node [midway,above,scale=0.7]{$\widehat{T}_{t\rightarrow i} = (R_i, t_i)$};
  \draw [>=stealth,->] (3,-0.5) |- (4,-1.6) |- (4.6,-1.8);
  \draw [>=stealth,->] (5.8,-2) -- (6.45,-2);
  \draw [>=stealth,->] (depth) -- (inverse warp);
  \draw [>=stealth,->] (inverse warp) -- (loss);

\pgfmathsetmacro{\cubex}{0.1}
\pgfmathsetmacro{\cubey}{1}
\pgfmathsetmacro{\cubez}{1}

\foreach \i in {1,...,3}
{
\pgfmathsetmacro{\cubey}{1.2-0.2*\i}
\pgfmathsetmacro{\cubez}{1.2-0.2*\i}
\draw[black,fill=white!30!red] (-2,-1.5,0) ++(0.2*\i,-0.1*\i,-0.1*\i) coordinate (a) -- ++(-\cubex,0,0) -- ++(0,-\cubey,0) -- ++(\cubex,0,0) -- cycle;
\draw[black,fill=white!10!red] (a) -- ++(0,0,-\cubez) -- ++(0,-\cubey,0) -- ++(0,0,\cubez) -- cycle;
\draw[black,fill=white!50!red] (a) -- ++(-\cubex,0,0) -- ++(0,0,-\cubez) -- ++(\cubex,0,0) -- cycle;
}

\foreach \i in {1,...,4}
{
\ifthenelse{\i = 2 \OR \i = 3}
  {\pgfmathsetmacro{\cubey}{0.8}
  \pgfmathsetmacro{\cubez}{0.8}}
  {\pgfmathsetmacro{\cubey}{1}
  \pgfmathsetmacro{\cubez}{1}}
;
\draw[black,fill=white!30!blue] (4.5,-1.5,0) ++(0.2*\i,0.5*\cubey-0.5,0.5*\cubez-0.5) coordinate (a) -- ++(-\cubex,0,0) -- ++(0,-\cubey,0) -- ++(\cubex,0,0) -- cycle;
\draw[black,fill=white!10!blue] (a) -- ++(0,0,-\cubez) -- ++(0,-\cubey,0) -- ++(0,0,\cubez) -- cycle;
\draw[black,fill=white!50!blue] (a) -- ++(-\cubex,0,0) -- ++(0,0,-\cubez) -- ++(\cubex,0,0) -- cycle;
}
  
  \end{tikzpicture}
\linebreak
\caption{General workflow architecture. \textit{target} and \textit{ref} indices ($t$ and $r$) are chosen randomly for each sequence (even within the same batch); output transformations of PoseNet are compensated so that $\widehat{T}_{t \rightarrow t}$ is identity. $D_0$ is a fixed nominal displacement.}
\label{fig:process}
\end{figure}

Inspired from both \cite{zhou2017unsupervised} and \cite{isprs-annals-IV-2-W3-67-2017}, we propose a framework for training a network similar to DepthNet, but from unlabeled video sequences. The approach can be decomposed into three tasks, as illustrated by Fig.~\ref{fig:process}:
\begin{itemize}
\item From a certain sequence of frames $(I_i)_{0 \leq i < N}$, randomly choose one target frame $I_t$ and one reference frame 
$I_r$, forming a pair to feed to DepthNet.
\item For each $i \in \left\llbracket 0,N \right\llbracket$ , estimate pose $\widehat{T}_{t\rightarrow i}$ = ($\widehat{R}_i, t_i$) of each frame $I_i$ relative to the target frame $I_t$, and compensate rotation of reference frame $I_r$ to $I^\textsc{stab}_r$ before feeding it to DepthNet, leading to the same situation as original DepthNet, fed with stabilized inputs in \cite{isprs-annals-IV-2-W3-67-2017}. As discussed in Section
\ref{sec:intro}, when considering UAV footage, rotation can be supervised.
\item Compute the depth map which for presentation purpose will be 
denoted
$\zeta$. DepthNet$(I^\textsc{stab}_r, I_t)$ = $\zeta(I^\textsc{stab}_r, I_t)$
\item 
Normalize the translation
to constrain it so that the displacement magnitude $t_r$ with respect to $I_r$, is always the same throughout the training. This point is very important, 
in order to guarantee the
equivariance between depth and motion, 
imposed by the original DepthNet training procedure.
\item As the problem is now made equivalent to the one used in \cite{zhou2017unsupervised}, perform a photometric reprojection of $I_t$ to every other frame $I_i$, thanks to depth $\zeta_t$ of $I_t$ and poses $\widehat{T}_{t \rightarrow i}$ computed before, and compute loss from photometric dissimilarity with $I_i$.
\end{itemize}

The whole reprojection process can be summarized by Eq. \ref{eq:repro}. where $K$ denotes the camera intrinsic, $p_t$ homogeneous coordinates of a pixel in frame $I_t$ and $\zeta_t(p_t)$ is the depth value of the pixel outputted by DepthNet. $p^i_t$ are homogeneous coordinates of that very pixel in frame $I_i$. To get the equivalent pixel in coordinate $p^i_t$, 
$I_i$ pixels are bilinearily interpolated.

Our algorithm, although relying on very little calibration, needs to get consistent focal length. This is due to the frame difference being dependent to focal length. However, this problem is easily avoided when training on sequences coming from the same camera. Also, as shown by Eq. \ref{eq:repro}, camera 
intrinsic matrix $K$ needs
to be known to compute warping and subsequent photometric reprojection loss properly. In practice, assuming optical center in the center of the focal plane worked for all our tests; this is corroborated by tests done by \cite{Mahjourian} where they used uncalibrated camera, only knowing approximate focal length.

\begin{equation}
\label{eq:repro}
\begin{array}{c}
\forall i \in \left\llbracket0,N\right\llbracket,p^i_t=K\widehat{T}_{t \rightarrow i}\left(\zeta_t(p_t)K^{-1}p_t\right) \\
\widehat{T}_{t \rightarrow i} \left\lbrace 
\begin{array}{lll}\mathbb{R}^3 & \rightarrow & \mathbb{R}^3 \\
X & \mapsto & \widehat{R}_{i}X + t_{i}
\end{array}\right.
\end{array}
\end{equation}

\subsection{Pose Estimation}
PoseNet, as initially introduced by Zhou \etal \cite{zhou2017unsupervised} is a classic fully convolutional neural Network that outputs 6 DoF transformation pose descriptor for each frame.
Output poses are initially relative to the last frame, and then compensated to be relative to the pose of the target frame. This way, PoseNet output is not dependent on the index of the target frame. Besides, computing by default with respect to the last frame makes the inference much more straightforward, as in real condition, target frame on which depth is computed should be the last of the sequence, to reduce latency.

\subsection{Frame stabilization}
In order to cancel rotation between target and reference frame, we can apply a warping using rotation estimation from PoseNet. When considering a transformation with no translation, Eq. \ref{eq:repro} no longer depends on the depth of each pixel, and becomes Eq. \ref{eq:stab}. As such, we can warp the frame to stabilize it using orientation estimation from PoseNet, before computing any depth.

\begin{equation}
\label{eq:stab}
p^r_t=KR_rK^{-1}p_t
\end{equation}

As mentioned in Section \ref{sec:intro}, UAV footages are either stabilized or with a reliable estimated orientation from inertial sensors. This information can be easily leveraged in our training workflow to supervise pose rotation, giving in the end only translation to estimate to PoseNet. In addition, when running in inference, no pose estimation is needed, and only DepthNet is used. A similar algorithm as Pinard \etal \cite{pinard:hal-01587658} can then be used to estimate absolute depth maps at a relatively low computational cost.

\subsection{Depth computing and pose normalization}
Thanks to the close relation between distance and optical flow for stabilized frames, \cite{isprs-annals-IV-2-W3-67-2017} proposed a network to compute depth from stabilized videos.
As depth is then provided assuming a constant displacement magnitude, the pose of the reference frame must be normalized to correspond to that magnitude.
As such, to get consistent poses throughout the whole sequence, we apply the same normalization ratio, as shown in Fig. \ref{fig:process}:
\begin{equation}
\forall i, t_i^{\textsc{norm}} = t_i\frac{D_0}{\epsilon + \Vert t_r\Vert}
\end{equation}
The main drawback of normalizing translations is the lack of guarantee about absolute output values. Since we only consider translations relatively to the reference, translations are estimated up to a scale factor that could be - when they are very large - leading to potential errors for rotation estimation, or - when they are very close to $0$ - leading to float overflow problems. To overcome these possible issues, along with classic $L_2$ regularization to avoid high values, we add a constant value $\epsilon$ to the denominator. The normalization is then valid only when 
$\epsilon \ll \left\Vert t_r \right\Vert$.

\subsection{Loss functions}
 Let us denote $\widehat{I}_i$ as the inverse warped image from $I_i$ to target image plane by $p^i_t$ and $\Vert\cdot\Vert_1$ 
the $L_1$ norm operator (corresponding here to the mean absolute value over the array). 
For readability, we contract $\zeta(I_r, I_t)$ into simply $\zeta$.
 
The optimization will then try to minimize the dissimilarities between the synthesized view $\widehat{I}_i$ and original frame $I_t$. As suggested by \cite{monodepth17}, raw pixel difference can be coupled with structural similarity (SSIM) \cite{wang2004image} maximization, in order to be robust to luminosity changes, either from camera auto-exposition or from non Lambertian surfaces. SSIM computation is detailed Eq. \ref{SSIM}, where $\mu$ and $\sigma$ are the local mean and variance operators, estimated by convolving  
the image with Gaussian kernels of size $3 \times 3$.
$C_1 = 0.01$ and $C_2 = 0.09$ are two constants.
Note that the use of convolution in SSIM increases the receptive field of the loss function with respect to the $L_1$ distance. 
\begin{equation}
\label{SSIM}
 \textsc{SSIM}(I_t, I_i) = \frac{(2\mu_{I_t}\mu_{I_i} + C_1) + (2 \sigma _{I_tI_i} + C_2)} 
    {(\mu_{I_t}^2 + \mu_{I_i}^2+C_1) (\sigma_{I_t}^2 + \sigma_{I_i}^2+C_2)}
\end{equation}
Our photometric loss $\mathcal{L}_p$ is then a mixture of the two, $\alpha$ being an empirical weight.

\begin{equation}
\mathcal{L}_p = \sum_i \Vert \widehat{I}_i - I_t\Vert_1 - \alpha \textsc{SSIM}(\widehat{I}_i, I_t)
\end{equation}

Along with frames dissimilarity, in order to avoid divergent depth values in occluded or low textured areas, we add a geometric smooth loss that tries to minimize depth relative Laplacian, weighted by image gradients. Also, contrary to single frame network, depthnet output $\zeta$ here is not normalized. Thus, we must scale it according to its mean value. The fraction here represents pixel wise division, $\nabla$ and $\Delta$ are the gradient and Laplacian operators respectively, obtained by $3\times3$ convolutions.

\begin{equation}
\mathcal{L}_g = \left\Vert\frac{|\Delta \zeta|}{\left\Vert\nabla I_t\right\Vert}\right\Vert_1 \times \frac{1}{\left\Vert \zeta \right\Vert_1}
\end{equation}


Finally, we apply this loss to multiple scales $s$ of DepthNet outputs, multiplied by a factor giving more importance to high resolution, and our final loss becomes

\begin{equation}
\mathcal{L} = \sum_s \frac{1}{2^s} \left(\mathcal{L}_p^s + \lambda\mathcal{L}_g^s \right)
\end{equation}

where $\lambda$ denotes an empirical weight.

\section{Experiments}

\begin{figure}
\label{images}
\begin{tabular}{cccc}
Input & Ground truth & Zhou \etal \cite{zhou2017unsupervised} & Ours\\
\includegraphics[scale=0.17]{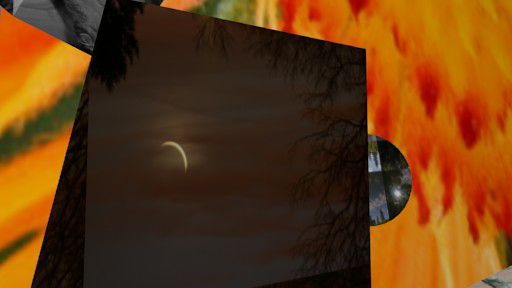} & \includegraphics[scale=0.17]{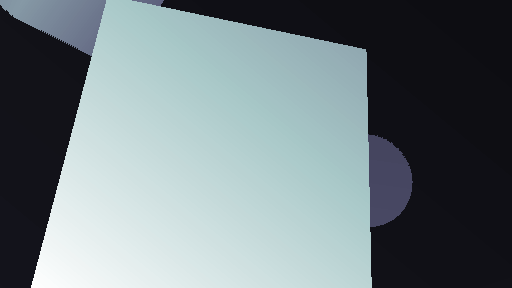} & \includegraphics[scale=0.17]{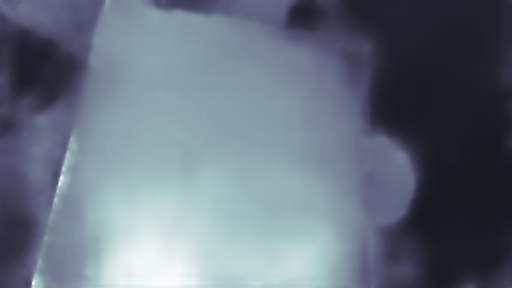} &
\includegraphics[scale=0.17]{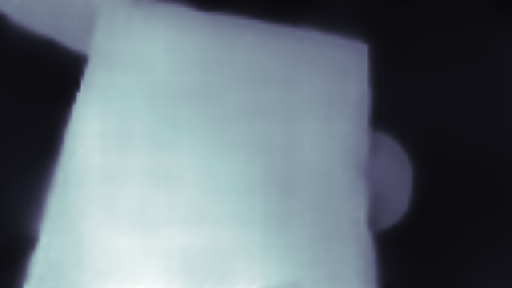}\\
\includegraphics[scale=0.17]{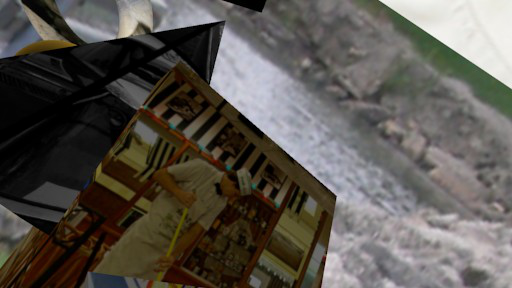} & \includegraphics[scale=0.17]{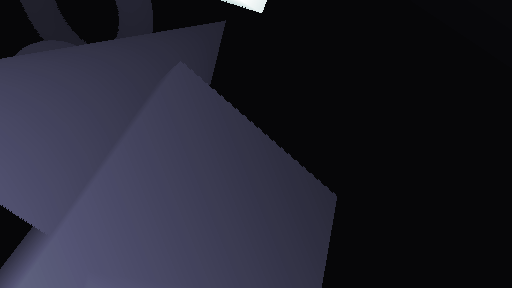} & \includegraphics[scale=0.17]{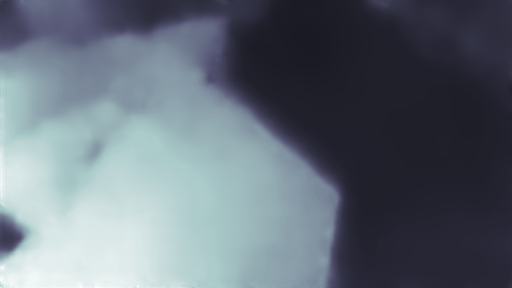} &
\includegraphics[scale=0.17]{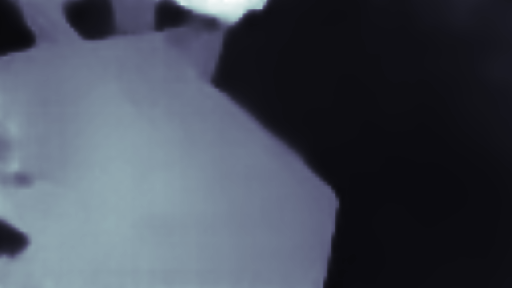} \\
\includegraphics[scale=0.17]{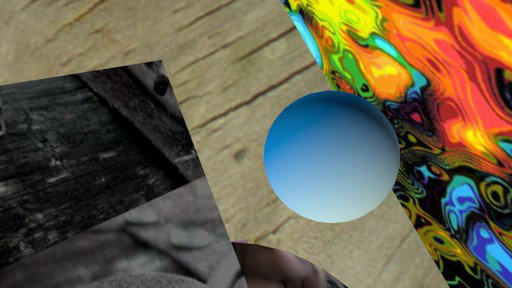} & \includegraphics[scale=0.17]{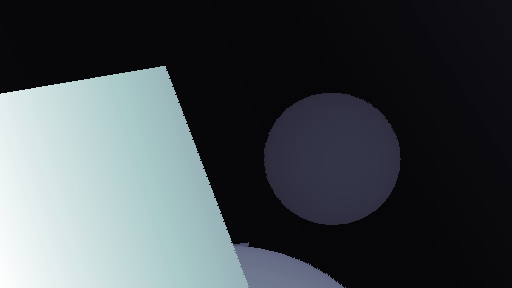} & \includegraphics[scale=0.17]{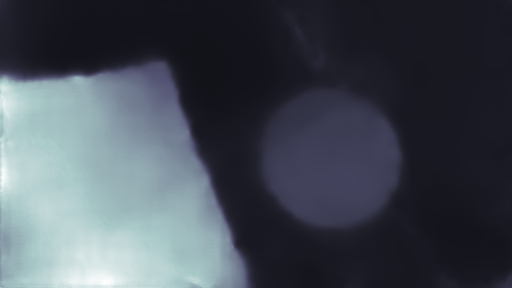} &
\includegraphics[scale=0.17]{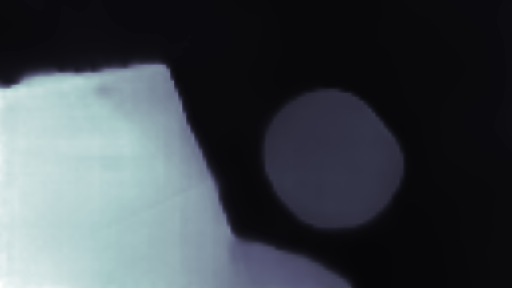} \\

\specialrule{.1em}{.05em}{.05em} 
\\
\includegraphics[scale=0.21]{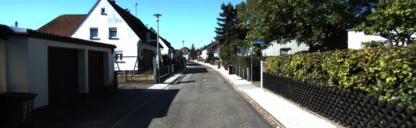} & \includegraphics[scale=0.21]{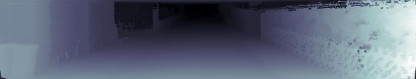} & \includegraphics[scale=0.21]{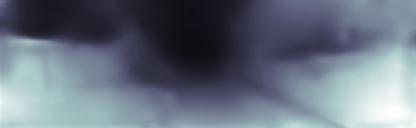} &
\includegraphics[scale=0.21]{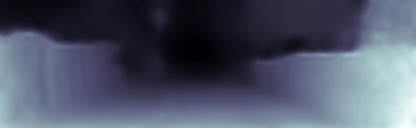} \\
\includegraphics[scale=0.21]{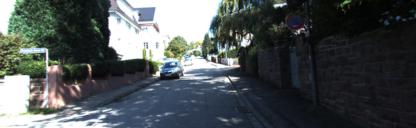} & \includegraphics[scale=0.21]{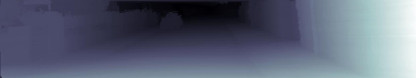} & \includegraphics[scale=0.21]{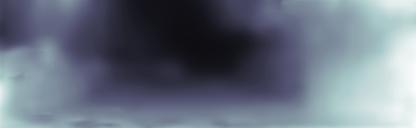} &
\includegraphics[scale=0.21]{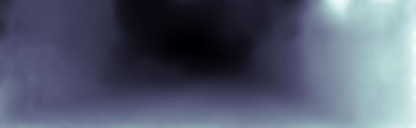} \\
\includegraphics[scale=0.21]{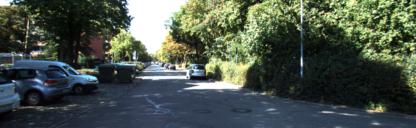} & \includegraphics[scale=0.21]{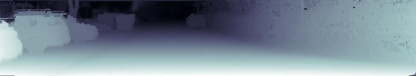} & \includegraphics[scale=0.21]{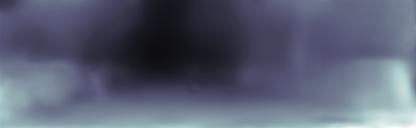} &
\includegraphics[scale=0.21]{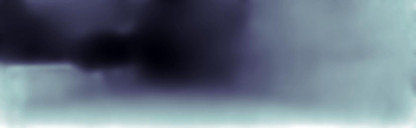} \\

\end{tabular}
\caption{
Comparison between our method and Zhou \etal \cite{zhou2017unsupervised} on updated Still Box \cite{isprs-annals-IV-2-W3-67-2017} and KITTI \cite{geiger2013vision}.
}
\end{figure}

\begin{figure}
\centering
\begin{tabular}{ccc}
Input & Zhou \etal \cite{zhou2017unsupervised} & Ours \\
\includegraphics[scale=0.28]{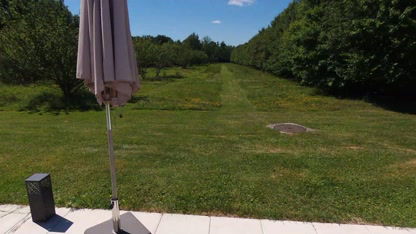} & \includegraphics[scale=0.28]{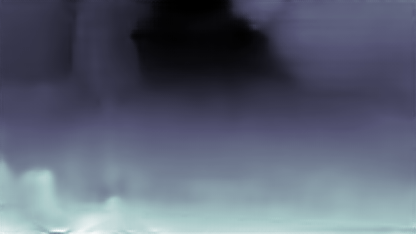} & \includegraphics[scale=0.28]{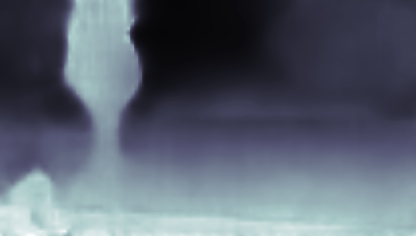} \\
\includegraphics[scale=0.28]{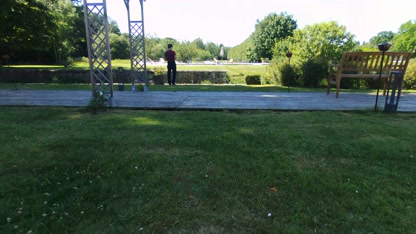} & \includegraphics[scale=0.28]{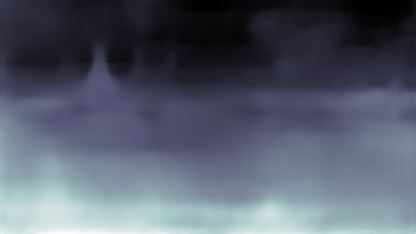} & \includegraphics[scale=0.28]{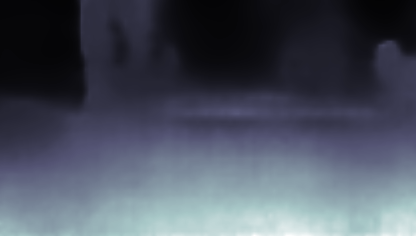} \\
\includegraphics[scale=0.28]{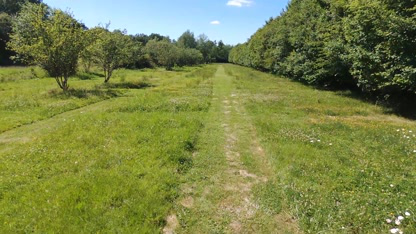} & \includegraphics[scale=0.28]{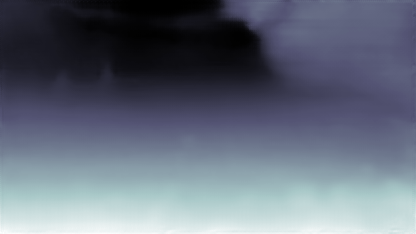} & \includegraphics[scale=0.28]{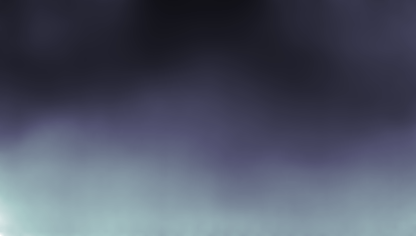} \\
\end{tabular}
\caption{\label{uav}
Subjective comparison of disparity maps between Zhou \cite{zhou2017unsupervised}  and our method on a small UAV dataset.}
\end{figure}

\begin{figure}
\centering
\begin{tabular}{cc}
\includegraphics[scale=0.45]{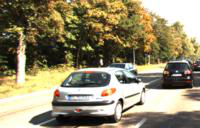} & \includegraphics[scale=0.45]{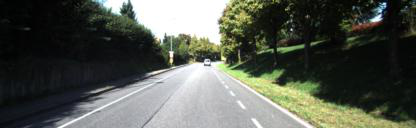}\\
\includegraphics[scale=0.45]{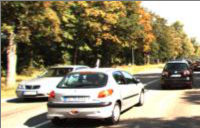} & \includegraphics[scale=0.45]{figure/KITTI/failure10.png}\\
\includegraphics[scale=0.45]{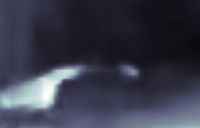} & \includegraphics[scale=0.45]{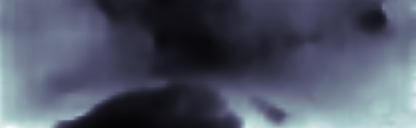}
\end{tabular}
\caption{\label{failures}
Some failure cases of our method on KITTI. First column is a detail of a larger image. The foreground car is moving forward and it's detected as far away, while the background car is moving toward us and is detected as close. Second column is a poorly textured road.
}
\end{figure}

\subsection{Training datasets}Our experiments were made on three different datasets. KITTI is one of the most well known datasets for training and evaluating algorithms on multiple computer vision tasks such as odometry, optical flow or disparity. It features stereo vision, LiDAR depth measures and GPS / RTK coupled with IMU for camera poses. During training we only used monocular frames and IMU values when supervising with orientation. We used LiDAR for evaluation. We applied the same training/validation/test split as \cite{zhou2017unsupervised}: about 40k frames for training and 4k for evaluation. We also discarded of the whole set scenes containing the 697 test frames from the Eigen \cite{eigen2014depth} split. We also constructed a filtered test set with the same frames as Eigen, but discarding 69 frames whose GPS position uncertainty was above $1$ m. This set was used when displacement data was needed.

We also conducted experiments on an updated version of Still Box, used in \cite{isprs-annals-IV-2-W3-67-2017}, in which we added random rotations
(\ie we draw an initial rotation speed that remains the same through the sequence). 
This dataset features synthetic rigid scenes, composed of basic 3d primitives (cubes, spheres, cones and tores) randomly textured using images scrapped from \textit{Flickr}. In this dataset, depth is difficult to infer from context, as shapes have random sizes and positions. Camera's movement is constrained to constant velocity (translation and rotation) throughout scenes of 20 pictures. 
The dataset contains 1500 training scenes and 100 test scenes, \ie 30k training frames and 2k validation frames.

Finally, we trained our network on a very small dataset of UAV videos, taken from the same camera the same day. We used a Bebop2 drone, with 30fps videos, and flew over a small area of about one hectare for 15 minutes.
The training set contains around 14k frames while the test set is a sequence of 400 frames. This dataset is not annotated, and only subjective evaluation can be done. 

\subsection{Implementation details}
DepthNet is almost the exact same as the one used in Pinard \etal \cite{isprs-annals-IV-2-W3-67-2017}. Its structure mainly consists of two components: the encoder and the decoder parts. The encoder follows the basic structure of VGG\cite{Simonyan14c}. The decoder is made up of deconvolution layers to bring back the spatial feature maps up to a fourth of the input resolution.
To preserve both high semantics and rich spatial information, we use skip connections between encoder and decoder parts at different corresponding resolutions. This is a multi-scale technique that was initially done in \cite{DFIB15}.
The main difference between DepthNet and our network is the ELU function \cite{clevert2015fast} applied to last depth output instead of identity.

PoseNet is the same as 
\cite{zhou2017unsupervised} which contains 8 convolutional with a stride of 2 layers followed by a global average pooling layer before final prediction. Every layer except the last one are post processed with ReLU activation \cite{nair2010rectified}.

We used PyTorch \cite{paszke2017automatic} for all tests and trainings, with empirical weights $\lambda = 3$ and $\alpha = 0.075$. We used Adam optimizer \cite{DBLP:journals/corr/KingmaB14} with learning rate of 
$2 \times 10^{-4}$
and $\beta_1 = 0.9$, $\beta_2 = 0.999$.

\subsection{Quality measurements and comparison with other algorithms}
In addition to 
using
standard measurements from \cite{eigen2014depth}, our goal is to measure how well a network would perform in real conditions. As stated in Section \ref{section:critic}, depth map scale factor must be determined from reasonable external data and not from explicit depth ground truth.

We thus compare our solution to \cite{zhou2017unsupervised} where the output is multiplied by the ratio between estimated displacement from PoseNet and actual values. For KITTI, displacement is determined by GPS RTK, but as we only need magnitude, speed from wheels would have been sufficient. When training was done with orientation supervision, we stabilized the frames before feeding them to DepthNet.
To test our method on our small UAV dataset, we first did a training on updated Still Box, then an unsupervised fine tuning. Likewise, when using Zhou \etal method \cite{zhou2017unsupervised}, we pretrained on KITTI before fine tuning on our video.
\subsection{Quantative training results}
\begin{table}
\resizebox{\columnwidth}{!}{
\begin{tabular}{|l|c|c|c|c|c|c|c|c|c|c|}
\hline
Method & \begin{tabular}{c}training\\set\end{tabular} & \begin{tabular}{c}scale\\factor\end{tabular} & supervision & \cellcolor{blue!15}Abs Rel & \cellcolor{blue!15}Sq Rel & \cellcolor{blue!15}RMSE & \cellcolor{blue!15}RMSE log & \cellcolor{red!25}$\delta < 1.25$ & \cellcolor{red!25}$\delta < 1.25^2$ & \cellcolor{red!25}$\delta < 1.25^3$ \\
\hline
Eigen \etal \cite{eigen2014depth} Coarse & K & GT & D & 0.214 & 1.605 & 6.563 & 0.292 & 0.673 & 0.884 & 0.957\\
Eigen \etal \cite{eigen2014depth} Fine & K & GT & D & 0.203 & 1.548 & 6.307 & 0.282 & 0.702 & 0.890 & 0.958\\
Zhou \etal \cite{zhou2017unsupervised} & K & GT & - & 0.183 & 1.595 & 6.709 & 0.270 & 0.734 & 0.902 & 0.959\\
Mahjourian \etal \cite{Mahjourian} & K & GT & - & 0.163 & 1.240 & 6.220 & 0.250 & 0.762 & 0.916 & 0.968 \\
Zhichao \etal \cite{yin2018geonet} & K & GT & - & 0.155 & 1.296 & 5.857 & 0.233 & 0.793 & 0.931 & \textbf{0.973} \\
Ranjan \etal \cite{DBLP:journals/corr/abs-1805-09806} & K & GT & - & \textbf{0.148} & \textbf{1.149} & \textbf{5.464} & \textbf{0.226} & \textbf{0.815} & \textbf{0.935} & \textbf{0.973} \\
\specialrule{.1em}{.05em}{.05em} 
Pinard \etal \cite{isprs-annals-IV-2-W3-67-2017} & S & P & D + O & 0.5071 & 7.1540 & 9.6209 & 0.5032 & 0.3960 & 0.6600 & 0.8138\\
\hline
Zhou \etal & K & P & - & 0.2786 & \textbf{2.7059} & 7.2956 & 0.3552 & 0.5816 & 0.8082 & 0.8982 \\
Ours & K & P & - & 0.3124 & 5.0302 & 8.4985 & 0.4095 & 0.5919 & 0.7961 & 0.8821 \\
Ours & S $\rightarrow$ K & P & D+O $\rightarrow$ - & 0.2940 & 3.9925 & 7.5727 & 0.3756 & 0.6092 & 0.8336 & 0.9090 \\
\hline
Ours & K & P & O & 0.2756 & 3.9335 & \textbf{7.2939} & 0.3539 & 0.6417 & 0.8457 & 0.9179 \\
Ours & S $\rightarrow$ K & P & D+O $\rightarrow$ O & \textbf{0.2706} & 4.4947 & 7.3119 & \textbf{0.3452} & \textbf{0.6778} & \textbf{0.8564} & \textbf{0.9242} \\

\hline
\end{tabular}
}
\caption{\label{table:KITTI}
Quantitative tests on KITTI\cite{geiger2013vision} Eigen split \cite{eigen2014depth}. Measures are the same as in Eigen \etal \cite{eigen2014depth}. For blue measures, lower is better, for red measures, higher is better. For training, K is the KITTI dataset \cite{geiger2013vision}, S is the Still Box dataset \cite{isprs-annals-IV-2-W3-67-2017}. For scale factor, GT is ground truth, P is pose. When scale was determined with pose, we discarded frames where GPS uncertainty was greater than $1m$. For supervision, D is depth and O is orientation. $\rightarrow$ denotes fine tuning.}
\end{table}

\begin{table}
\resizebox{\columnwidth}{!}{
\begin{tabular}{|l|c|c|c|c|c|c|c|c|c|}
\hline
Method & \begin{tabular}{c}scale\\factor\end{tabular}& supervision & \cellcolor{blue!15}Abs Rel & \cellcolor{blue!15}Sq Rel & \cellcolor{blue!15}RMSE & \cellcolor{blue!15}RMSE log & \cellcolor{red!25}$\delta < 1.25$ & \cellcolor{red!25}$\delta < 1.25^2$ & \cellcolor{red!25}$\delta < 1.25^3$ \\
\hline
Pinard \etal \cite{isprs-annals-IV-2-W3-67-2017} & P & D + O & \textbf{0.2120} & \textbf{2.0644} & \textbf{7.0669} & \textbf{0.2959} & \textbf{0.7091} & \textbf{0.8810} & \textbf{0.9460}\\
\hline
Zhou \etal \cite{zhou2017unsupervised} & GT & - & 0.5005 & 11.4189 & 15.7207 & 0.6012 & 0.4969 & 0.6767 & 0.7671 \\
Zhou \etal \cite{zhou2017unsupervised} & P & - & 0.8109 & 11.9956 & 17.2740 & 0.6928 & 0.3475 & 0.5733 & 0.7136 \\
\hline
Ours & P & - & 0.4684 & 10.9247 & 15.7560 & 0.5440 & 0.4524 & 0.6772 & 0.8037 \\
Ours & P & O & \textbf{0.2970} & \textbf{5.2827} & \textbf{10.5090} & \textbf{0.4041} & \textbf{0.6684} & \textbf{0.8405} & \textbf{0.9058} \\
\hline
\end{tabular}
}
\caption{\label{table:StillBox} 
Quantitative tests on StillBox, no pretraining has been done. The supervised Pinard \etal \cite{isprs-annals-IV-2-W3-67-2017} method is here to give an hint on a theoritical limit since it uses the same network, but with depth supervision}
\end{table}
Table \ref{table:KITTI} presents quantitative results compared to prior works. We tried 5 different versions of our network. The first one is the exact same as \cite{isprs-annals-IV-2-W3-67-2017}, only trained on StillBox. It serves as a baseline purpose, without finetuning. The other four configurations are training from scratch or finetuning from StillBox, and training with orientation supervision or not.

As we might expect, on KITTI our method fails to converge as well as single image methods using classic relative depth quality measurement. However, when scale factor is determined from poses, we match the performance of the adapted method from \cite{zhou2017unsupervised}. It can also be noted that finetuning provides a better starting point for our network, and that when available on a training set, orientation supervision is very advantageous.

When trying to train a Depth network with stabilized videos, it is then strongly recommended to do a first supervised training on a synthetic dataset such as StillBox.

Some failed test cases can be seen on Fig.\ref{failures}. The main sources of error are moving objects and poorly textured
areas (especially concrete roads), even though we applied depth smooth geometric loss. Our attempt at explaining this failure is the large optical flow value compared to low textured area, meaning matching spatial structures is difficult. However, as KITTI acquisition rate is only 10 fps, we believe this problem would be less common on regular cameras, with typical rates of 30 fps or higher.

Table \ref{table:StillBox} presents results on the updated Still Box dataset. Zhou \etal \cite{zhou2017unsupervised} performs surprisingly well given the theoretical lack of visual context in this dataset. However, our method performs better, whether from orientation supervision or completely unsupervised. We also compare it to supervised DepthNet from \cite{isprs-annals-IV-2-W3-67-2017}. This can be considered a theoretical limit for training DepthNet on Still Box since 
\cite{isprs-annals-IV-2-W3-67-2017}
is completely supervised, and our training method is very close to it, indicating the good convergence of our model and thus our training algorithm and loss design validity.

\subsection{Domain adaptation results}
Finally, Fig.\ref{uav} (left) compares Zhou \etal \cite{zhou2017unsupervised} and our method on some test frames of our small UAV dataset. Our methods shows much better domain adaptation when fine tuning in a few-shot learning fashion. Especially for foreground objects, as Zhou \etal \cite{zhou2017unsupervised} blends it with the trees near the horizon, which is very problematic for navigation. A video with result comparison is 
provided as supplementary material to this paper.

Table \ref{table:KITTI_reversed} compares domain robustness without any training: we tried inference on 
an upside-down
KITTI test set, with ground up and sky down, and our method performs much better than Zhou \etal \cite{zhou2017unsupervised}, which is completely lost and performs worse than 
inferring 
a constant plane. However, our method is not as performing as Pinard \etal \cite{isprs-annals-IV-2-W3-67-2017} which score was expected to be the same as in Table \ref{table:KITTI}, since it has not been trained on any KITTI frames, whether regular or reversed. This shows that our network may have learned to infer depth from both motion and context, which can be considered as a compromise between our two competitors, Zhou \etal \cite{zhou2017unsupervised} relying only on context and Pinard \etal \cite{isprs-annals-IV-2-W3-67-2017} only on motion.

\begin{table}
\resizebox{\columnwidth}{!}{
\begin{tabular}{|l|c|c|c|c|c|c|c|c|c|c|}
\hline
Method & \begin{tabular}{c}training\\set\end{tabular} & \begin{tabular}{c}scale\\factor\end{tabular}& supervision & \cellcolor{blue!15}Abs Rel & \cellcolor{blue!15}Sq Rel & \cellcolor{blue!15}RMSE & \cellcolor{blue!15}RMSE log & \cellcolor{red!25}$\delta < 1.25$ & \cellcolor{red!25}$\delta < 1.25^2$ & \cellcolor{red!25}$\delta < 1.25^3$ \\
\hline
Pinard \etal \cite{isprs-annals-IV-2-W3-67-2017} & S & P & D + O & \textbf{0.4622} & \textbf{6.0229} & \textbf{9.2277} & \textbf{0.4807} & \textbf{0.4149} & \textbf{0.6863} & \textbf{0.8349}\\
\hline
Constant Plane & - & GT & - & 0.4568 & 4.8516 & 12.0848 & 0.6000 & 0.2962 & 0.5488 & 0.7524\\
Zhou \etal \cite{zhou2017unsupervised} & K & GT & - & 0.5931 & 7.5410 & 12.9943 & 0.7340 & 0.2223 & 0.4342 & 0.6263 \\
Zhou \etal \cite{zhou2017unsupervised} & K & P & - & 1.5879 & 62.1068 & 21.1424 & 0.9579 & 0.1688 & 0.3260 & 0.4744 \\
\hline
Ours & S $\rightarrow$ K & P & - & \textbf{0.6484} & \textbf{15.3906} & \textbf{12.4324} & 0.6245 & 0.3820 & 0.6168 & 0.7607 \\
Ours & S $\rightarrow$ K & P & O & 0.7158 & 18.8145 & 12.5424 & \textbf{0.5987} & \textbf{0.4024} & \textbf{0.6370} & \textbf{0.7723} \\
\hline
\end{tabular}
}
\caption{\label{table:KITTI_reversed}
Quantitative tests on 
upside-down
KITTI \cite{geiger2013vision}:
sky is down and ground is up. No training has been done. Constant Plane outputs the same depth for every pixel}
\end{table}

\section{Conclusion and future work}

We 
have
presented a novel method for unsupervised depth learning, using not only depth from context but also from motion. This method leverages the context of stabilized videos, which is a midway between 
common
use cases of stereo rigs and unconstrained ego-motion. As such, our algorithm provides a solution with embedded deployment in mind, especially for UAVs navigation, and requires only video and inertial data to be used.

Our method is also much more robust to domain changes, which is an important issue when dealing with deployment in large scale consumer electronics on which it is impossible to predict all possible contexts and situations, and our method outperforms single frame systems on unusual scenes.

The greatest limitation of our algorithm is the necessity of rigid scenes \emph{even in inference}. Indeed, for single frame depth estimation or stereo, rigidity is not necessary during training. If an object moves between two frames, depth errors may compensate over the dataset if the object’s movements are evenly distributed. As a consequence, some datasets, although usable for training, are not particularly suited for quality measurement since some scenes are not rigid. This is a strong limitation and one of our goals for a future work.

\bibliographystyle{splncs}

\bibliography{egbib}
\end{document}